%% file: root.tex
\documentclass[letterpaper, 10 pt, conference]{ieeeconf} 

\IEEEoverridecommandlockouts

\usepackage{graphicx} 
\usepackage[utf8]{inputenc}
\usepackage{todonotes} 
\usepackage{amsmath,amsfonts,booktabs,cite} 
\usepackage{siunitx,textcomp} 
\usepackage{algorithmic,algorithm} 
\usepackage[hidelinks]{hyperref} 
\usepackage{multicol}
\usepackage{epstopdf} 
\usepackage{subcaption}
\let\labelindent\relax
\usepackage[inline]{enumitem} 
\usepackage[nolist]{acronym} 
\usepackage{colortbl} 	    
\input{acronyms.tex}

\usepackage{standalone,tikz,pgfplots} 
\pgfplotsset{compat=1.13}

\input{macro.tex} 

\title{\LARGE \bf%
Robust Grasp Planning Over Uncertain Shape Completions}

\author{Jens~Lundell, Francesco~Verdoja 
and Ville~Kyrki%
\thanks{This work was supported by the Strategic Research Council at Academy of 
Finland, decision 314180. We gratefully acknowledge the support of NVIDIA 
Corporation with the donation of the Titan Xp GPU used for this research.}
\thanks{J.~Lundell, F.~Verdoja and V.~Kyrki are with School of Electrical 
Engineering, Aalto University, Finland. 
\texttt{\{firstname.lastname\}{@}aalto.fi}}}

\begin{document}

\maketitle
\thispagestyle{empty}
\pagestyle{empty}


\begin{abstract}
We present a method for planning robust grasps over uncertain shape 
completed objects. For shape completion, a deep neural network is trained to 
take a partial view of the object as input and outputs the completed shape as a 
voxel grid. The key part of the network is dropout layers which are enabled not 
only during training but also at run-time to generate a set of shape samples 
representing the shape uncertainty through Monte Carlo sampling. Given the set 
of shape completed objects, we generate grasp candidates on the mean object 
shape but evaluate them based on their joint performance in terms of analytical 
grasp metrics on all the shape candidates. We experimentally validate and 
benchmark our method against another \sota{} method with a \barrett{} on 
90000 grasps in simulation and 200 grasps on a real \panda{}. All experimental 
results show statistically significant improvements both in terms of grasp 
quality metrics and grasp success rate, demonstrating that planning 
shape-uncertainty-aware grasps brings significant advantages over solely 
planning on a single shape estimate, especially when dealing with complex or 
unknown objects.
\end{abstract}


\input{sections/intro.tex}
\input{sections/related.tex}

\input{sections/method.tex}

\input{sections/exp.tex}
%
\input{sections/concl.tex}


%



\bibliographystyle{IEEEtran}
\bibliography{refs}

\end{document}

%% file: acronyms.tex
\newacro{bn}[BN]{Batch Normalization}
\newacro{relu}[ReLU]{Rectified Linear Unit}
\newacro{adam}[Adam]{Adaptive Moment Estimation}
\newacro{ai}[AI]{Artificial Intelligence}
\newacro{dl}[DL]{Deep Learning}
\newacro{dnn}[DNN]{Deep Neural Network}
\newacro{cnn}[CNN]{Convolutional Neural Network}
\newacro{bnn}[BNN]{Bayesian Neural Network}
\newacro{mc}[MC]{Monte-Carlo}
\newacro{dof}[DOF]{Degrees-Of-Freedom}
\newacro{ods}[USN]{Uncertain Shape Network}
\newacro{od}[SN]{Shape Network}
\newacro{gp}[GP]{Gaussian Processes}
\newacro{mcmc}[MCMC]{Markov Chain Mote Carlo}
\newacro{psdf}[p-SDF]{probabilistic Signed Distance Function}
\newacro{gpis}[GPISs]{Gaussian Process Implicit Surfaces}
\newacro{va}[V]{Varley}

%% file: macro.tex
\newcommand{\figref}[1]{\hyperref[#1]{Figure~\ref*{#1}}}
\newcommand{\tabref}[1]{\hyperref[#1]{Table~\ref*{#1}}}
\newcommand{\secref}[1]{\hyperref[#1]{Section~\ref*{#1}}}
\newcommand{\algoref}[1]{\hyperref[#1]{Algorithm~\ref*{#1}}}
\newcommand{\figsref}[2]{Figures~\ref{#1}-\ref{#2}}
\newcommand{\subfigref}[1]{(\subref{#1})}
\newcommand{\ra}[1]{\renewcommand{\arraystretch}{#1}}
\newcommand{\tbs}[1]{\renewcommand{\tabcolsep}{#1pt}}
\newlength\myindent
\newcommand{\matr}[1]{\mathbf{#1}}
\DeclareMathOperator*{\argmax}{arg\,max}

\newcommand{\de}[1]{\operatorname{d}\!#1}

\def\panda{\textit{Franka Emika Panda}}
\def\barrett{\textit{Barrett hand}}
\def\sota{state-of-the-art}

\def\ie{\textit{i.e.,}}
\def\eg{\textit{e.g.,}}
\def\etal{\textit{et al.}}
\def\graspit{GraspIt!}
\def\pc{point-cloud}

\def\epst{\multicolumn{1}{c}{$\epsilon$}}
\def\vt{\multicolumn{1}{c}{$v$}}
\def\nat{\multicolumn{1}{c}{--}}

\definecolor{significant}{RGB}{217,217,217}
\definecolor{verysignificant}{RGB}{255,255,255}

%% file: sections/intro.tex
\section{Introduction}
\label{sec:intro}

In robotic grasping, knowing the object shape allows for better grasp planning. 
However, in many environments it is impossible to know a priori the shape of 
all possible objects. For this reason, the object to be grasped is usually 
perceived through some sensory input, commonly vision. However, one of the 
main problems with this approach is that only one side of the object is perceived, due to object self-occluding its back side. To cope with 
this limitation, essentially two options exist: 
\begin{enumerate*}[label=(\roman*)]
	\item use the information perceived and generate grasps based on this 
	knowledge alone \cite{mahler2017dex,schmidt_grasping_2018}, or 
	\item from the same input extract additional knowledge of the object with, 
	for example, semantic segmentation \cite{schwarz2018rgb} or shape 
	completion \cite{varley_shape_2017,mahler2015gp} and plan grasps accordingly.
\end{enumerate*}

In this paper, we focus on the latter, that is shape completion, and train a 
deep network to estimate the complete object shape. However, in contrast to 
most recent work in the field 
\cite{han_high-resolution_2017,dai2017complete,yang20173d} where 
the focus is explicitly on generating more exact point estimates of the shape, 
this work takes another viewpoint of the problem by also modeling the 
uncertainty over the completed shape. This uncertainty can then be incorporated 
into probabilistic grasp planners to enable robust grasp planning over 
uncertain shapes.

\begin{figure}
	\centering
	\includegraphics[width=.8\linewidth]{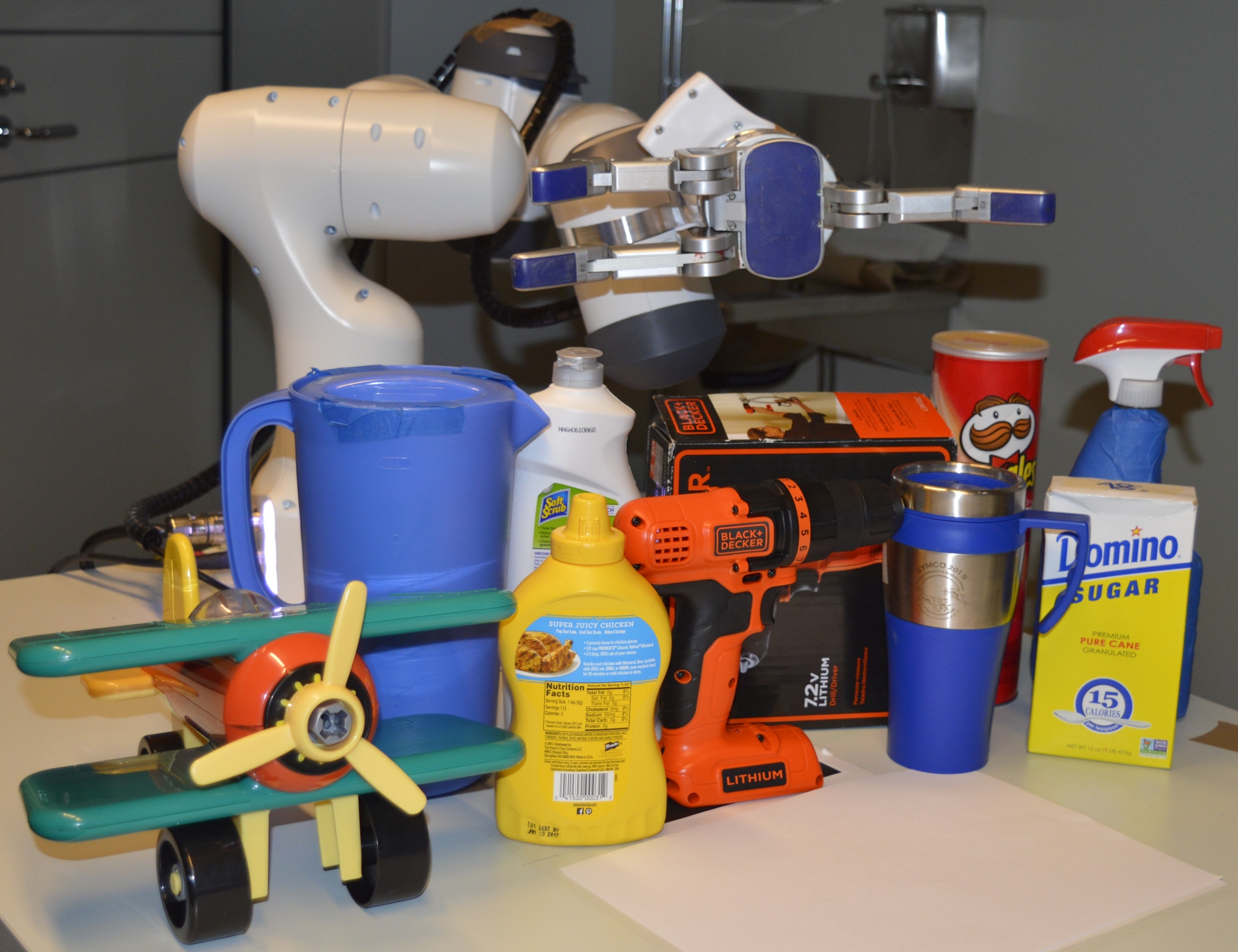}
	\caption{\label{fig:real_experiment}The real world setup with the \panda{} 
	robot and the 10 objects used in the experiments.}
\end{figure}


To this end, we propose a \ac{dnn} architecture with dropout layers active both 
during training and testing (\secref{sec:SBSC}). With such a structure, 
uncertainty is quantified from the difference in samples generated from feeding 
the same input through the network multiple times but each time with a 
different dropout mask. This use of dropout was originally proposed in 
\cite{gal_dropout_2016} as a method for approximate 
Bayesian inference in deep Gaussian processes. 

Another open issue is how to plan grasps over uncertain shapes. We address this by incorporating the uncertainty into probabilistic 
grasp planning (\secref{sec:PGP}) and propose a computationally tractable 
method for planning robust grasps over uncertain shapes (\secref{sec:RobustGraspPlanning}). The proposed method is experimentally validated (\secref{sec:exp}) by comparing it to 
a deep learning-based method \cite{varley_shape_2017} that only 
generates a point estimate of the shape in terms of shape reconstruction (\secref{sec:GCR}), analytical grasp quality metrics in simulation 
(\secref{sec:GSim}), and grasp success rate on the real \panda{} seen in 
\figref{fig:real_experiment} (\secref{sec:GReal}). Simulations of 90000 grasps demonstrate a statistically significant improvement in recognizing grasps with high quality metrics when including the shape uncertainty. The physical experiments of 200 grasps also show statistically significant improvement in terms of higher grasp success rate when planning is performed over the shape distribution compared to solely planning on a point estimate of the shape.

The main contributions of this work are:
\begin{enumerate*}[label=(\roman*)]
	\item a novel shape completion \ac{dnn} architecture able to capture shape 
	uncertainties, 
	\item a probabilistic grasp planning method that utilizes the shape 
	uncertainty to propose robust grasps, and 
	\item an empirical evaluation of the proposed method against \sota{}, 
	presenting, both in simulation and on real hardware, a statistically 
	significant improvement using the proposed method both in terms of grasp 
	ranking and on grasp success rate.
\end{enumerate*}

%% file: sections/related.tex
\section{Related Works}
\label{sec:related}

\subsection{Probabilistic Grasp Planning}

Probabilistic grasp planning addresses the issue of planning grasps under 
uncertainty. Typical uncertainties in robotic grasping are related to object 
pose uncertainty \cite{hsiao2011bayesian, 
laaksonen2012probabilistic,kim_physically_2013}, 
object shape uncertainty 
\cite{li2016dexterous,mahler2015gp,chen2018probabilistic}, or friction and 
contact position uncertainty \cite{zheng_coping_2005}.

For instance, Hsiao \etal{} \cite{hsiao2011bayesian} developed a Bayesian 
framework to generate grasps that were robust to both pose and shape uncertainty 
as well as robot motion error  by simulating grasps on deterministic mesh and 
point cloud models. A similar framework was also used in 
\cite{laaksonen2012probabilistic}, where they use \ac{gp} to model the 
grasp stability from tactile feedback and \ac{mcmc} to propose stable grasps.

In terms of grasping under shape uncertainty, Li \etal{} \cite{li2016dexterous} 
proposed a method that models the shape uncertainty with a \ac{gp}, encodes it 
as a grasp planning constraint, and optimizes for a grasp that minimizes the 
distance  between the center of the contact points and the origin of the 
object. They then cast the problem of computing a hand configuration that can 
realize the grasping location as a learning problem. However, the shape 
reconstruction performance of that method is conditioned on possibility to 
sample any subset of point of a complete \pc~of the object. This limitations 
makes it difficult to apply on unknown objects, something we address in this 
work.

Mahler \etal{} \cite{mahler2015gp} instead use \ac{gpis} to represent shape 
uncertainty and measures grasp quality by the probability of force closure. 
Tangential to the works using \ac{gp} to model shape uncertainty is the work in 
\cite{chen2018probabilistic}, where they use \ac{psdf} to represent shape 
uncertainty and a simulated annealing approach to search and optimize grasps. 
However, \cite{mahler2015gp} only consider shape uncertainty in 2D and 
\cite{chen2018probabilistic} requires multiple views of objects for shape 
reconstruction, whereas our method reasons about the shape uncertainty in 3D 
from only one viewpoint. 

\subsection{Shape Completion}
In this work we refer to shape completion methods 
as shape reconstruction from an incomplete \pc. Most such methods fall into one 
of three distinct categories: 1) Geometric approaches, 2) Template-based 
approaches, and 3) Deep learning-based approaches. Here, we will mainly focus 
on the third category and refer to \cite{Berger2014StateOT} for an in depth 
survey over the first two.

The first category, geometric approaches, includes symmetry driven 
\cite{kazhdan2006poisson} and heuristic methods \cite{schnabel2009completion}. 
The former reconstructs a shape by mirroring the input object through its 
symmetry axis while the latter reconstructs the shape by combining primitives 
such as planes and cylinders into one final shape. Template-based approaches, 
on the other hand, seek to match the input to an object in a database and then 
deform it to match the input \cite{pauly2005example}. Although many of these 
approaches originates from the computer vision perspective where the focus is 
on achieving better shape reconstruction, similar work have also been 
successfully applied in robotics. For example to facilitate robotic grasping 
by using symmetry 
\cite{bohg_mind_2011}, heuristics 
\cite{miller_automatic_2003}, or template matching methods
\cite{rodriguez_transferring_2018}. However, these methods are only applicable 
for specific sets of objects. For example, mirroring fails if the object has 
more than one axis of symmetry, whereas heuristics and template based matching 
are computationally restricted to specific subset of objects. Our method, on 
the other hand, do not rely on either symmetry or a known set of objects and is 
therefore more general.

More modern shape completion methods are based upon deep learning 
\cite{varley_shape_2017,varley2018multi,han_high-resolution_2017,dai2017complete,yang20173d}.
 In this context, most recent improvements originate from more refined network 
structures \cite{han_high-resolution_2017,yang20173d}, from the inclusion of 
semantic object classification \cite{dai2017complete}, or from
incorporating other sensing modalities such as tactile information 
\cite{varley2018multi}.

In terms of similar methods applied to robotics, only two works exist 
\cite{varley_shape_2017} and \cite{varley2018multi}. Both of these work used 
shape completion to facilitate robotic grasping where the latter one extended 
the former by incorporating tactile information of the object to improve shape 
reconstruction. In \cite{varley_shape_2017} the authors propose an hourglass 
\ac{cnn} architecture to reconstruct the shape given a voxel grid of the input 
\pc{}. That architecture, however, employs fully connected up-sampling layers 
resulting in a network with approximately 300 million parameters. Our network, 
on the other hand, has approximately 10 times less parameters as it uses 
convolutional layers throughout.

Together, all deep learning-based shape completion methods have solely focused 
on improving quality of the object shape estimate. This work, on the other 
hand, shifts the focus from single point estimates and explores estimation of 
the shape uncertainty. This is especially valuable in robotic grasping as it 
allows planning grasps that are robust to shape uncertainty.

%% file: sections/method.tex
\section{Method}
\label{sec:method}

\subsection{Probabilistic Grasp Planning}
\label{sec:PGP}
Let us define $G$ as the set of all possible grasps, represented as 6D 
end-effector pose and joint values, obtained by a grasp planning strategy.
Traditional grasp planning can be formalized in a probabilistic framework as an 
attempt at generating a candidate grasp $g\in G$ whose stability $S$ is 
maximized over a perfectly known object shape $o$. Formally,
\begin{align}
\label{eq:tradGrasp}
\argmax_{g \in G}P(S \mid g,o)\enspace,
\end{align}
where $P(S \mid g,o)$ is usually estimated by using some defined grasp quality 
metric such as the epsilon- ($\epsilon$-) or volume-measure ($v$-measure) 
\cite{miller1999examples}. 

In this work we follow the same procedure of maximizing a grasp quality metric 
but do not assume prefect knowledge of the object shape. Instead, the shape is 
modeled as a probability distribution $P(O \mid r)$ conditioned on some
measurements $r$ representing, for example, a partial view of the object in the 
form of a \pc. Consequently, we have that
\begin{align}
\label{eq:probGrasping}
P(S \mid G, r)=\int P(S \mid G,O)P(O \mid r) \de{O}\enspace.
\end{align}

The marginalization over shapes $O$ in \eqref{eq:probGrasping} is intractable 
beyond the simplest cases where we only target a specific class of objects 
\cite{hsiao2011bayesian}. To circumvent this problem, we propose using a 
sampling scheme to approximate \eqref{eq:probGrasping} 
\begin{align}
\label{eq:samplingEQ}
P(S \mid G, r)\approx \frac{1}{N}\sum_{i=1}^{N} P(S \mid G, o_i)\enspace,
\end{align}
where $o_i \sim P(O \mid r)$. The actual sampling process $o_i \sim P(O \mid 
r)$ is described in \secref{sec:SBSC}. 

Maximizing \eqref{eq:samplingEQ} requires a set of grasps candidates $g \in G$. However, generating those on all shapes $o_i$ is 
computationally infeasible. For that reason, instead of planning separate 
grasps on each shape $o_i$, we compute a mean shape 
$\hat{o}=\mathbb{E}[o_i]$ and only sample grasps on that, obtaining a set of 
candidate grasps $\hat{G} \subseteq G$. Finally, each grasp candidate in $\hat{G}$ is 
evaluated on all samples $o_i$ and the one with the highest average grasp 
quality metric across all samples is considered most robust. Formally, the 
most robust grasp solves
\begin{align}
\label{eq:graspMax}
\argmax_{g\in G}P(S \mid g, r) \approx \argmax_{\hat{g} \in \hat{G}} 
\frac{1}{N}\sum_{i=1}^{N}P(S \mid \hat{g}, o_i)\enspace.
\end{align}


\subsection{Sampling Based Shape Completion}
\label{sec:SBSC}
\begin{figure*}[]
	\centering
	\includegraphics[width=\textwidth]{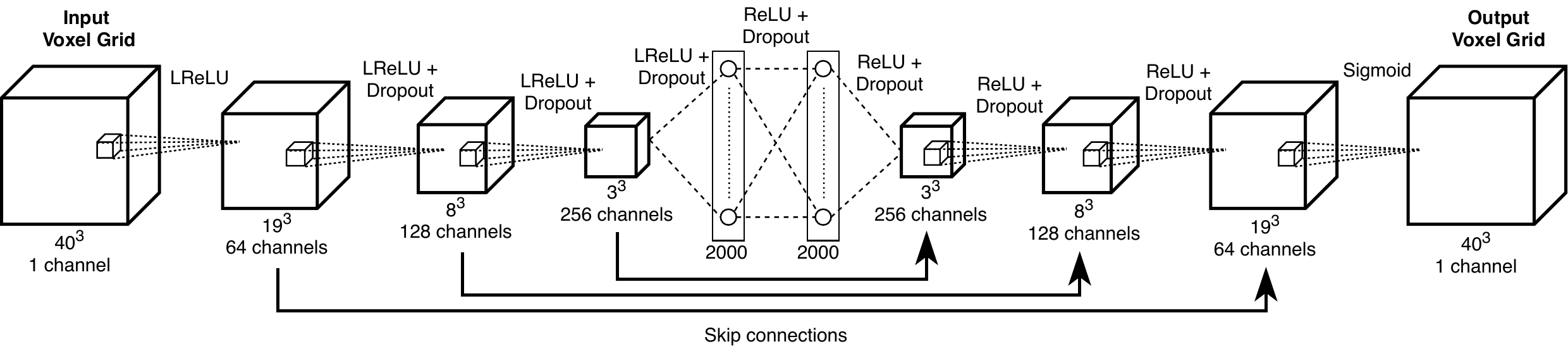}
	\caption{\label{fig:NN_Architecture}The proposed network architecture. In this architecture, ReLU stands for rectified linear unit and LReLU for leaky rectified linear unit.}
\end{figure*}
One of the crucial parts of the framework presented in \secref{sec:PGP} is the 
sampling process we employ to estimate the posterior distribution $P(O \mid 
r)$, \ie{} how to obtain a distribution of shapes from a partial sensor 
reading. One options is to use a \ac{bnn}, but for most forms of neural 
networks, such as the one used in this work, computing the full posterior is 
computationally intractable \cite{snoek2015scalable}. Therefore, we propose to 
approximate $P(O \mid r)$ using variational inference through the use of 
dropout sampling \cite{gal_dropout_2016}, where samples generated by having the 
dropout layers active also during test-time and feeding the same input through 
the network multiple times are used to approximate the full posterior $P(O \mid 
r)$. This procedure, known as \ac{mc}-Dropout, is a method to achieve approximate inference in Gaussian processes and \acp{dnn} \cite{gal_dropout_2016}.

To enable using \ac{mc}-Dropout, and in turn estimate $P(O \mid r)$, we propose to use the \ac{dnn} 
$\mathcal{H}$ shown in \figref{fig:NN_Architecture} to generate a shape $o$ 
given a sensor reading $r$, that is $o=\mathcal{H}(r)$. The network 
architecture is inspired from \cite{dai2017shape} but with a few important 
modifications to tailor it to our application: the 
input data dimensions (\ie{} voxel size) is changed ($40^3$ compared to $32^3$), to have 
fewer layers, and most importantly seven dropout layers \cite{srivastava2014dropout} are included. The network is trained in a supervised 
fashion with the cross-entropy error \cite{varley_shape_2017}
\begin{align*}
E(o,\tilde{o})=-(o\log(\tilde{o})+(1-o)\log(1-\tilde{o})),
\end{align*}
where $o$ is the network output and $\tilde{o}$ is the ground-truth target. 

Shape samples $o_i \sim P(O \mid r)$ for one measurement $r$ are then
generated with \ac{mc}-Dropout.
The results is a set $O_{\mathcal{I}} = \{o_i\}_{i=1}^N$ of shapes. Given this 
set, we then evaluate the mean shape $\hat{o}$ and use it to generate the 
subset of grasps candidates $\hat{G}$ as mentioned in \secref{sec:PGP}.

\subsection{Robust Grasp Planning Over Uncertain Shapes}
\label{sec:RobustGraspPlanning}

We propose \algoref{alg:grasp_planning} to plan robust grasps over uncertain 
shapes. In short, the algorithm first create a number of sample shapes based on 
an observation (lines \ref{op:dropoutsamplesbegin}-\ref{op:dropoutsamplesend}), 
then plans a set of candidate grasps on the mean shape (line 
\ref{op:graspPlanning}), and finally evaluate each of the grasp candidates on the 
entire set of shapes (lines 
\ref{op:robustPlanningBegin}-\ref{op:robustPlanningEnd}).

To generate the shape candidates, the algorithm first voxelizes the input \pc{} (line \ref{op:voxelize}). Then, $I$ samples are generated by following the procedure detailed in \secref{sec:SBSC}. To transform a voxel grid the network outputs into a mesh we used the algorithm $\textsc{Shape Completion}$ proposed in \cite{varley_shape_2017}.

The mean object mesh is created by first averaging all voxel grids into a mean voxel grid (line \ref{op:meanvoxel}) and then transforming it into a mesh (line \ref{op:meanmesh}). Grasps are then planned on the mean mesh (line \ref{op:graspPlanning}) and a procedure to do this is described in \secref{sec:exp_setup}.

Next, each grasp is separately evaluated on every sample (line 
\ref{op:evalGrasp}). Finally, given the grasp quality metrics on all samples, 
grasps are ranked (line \ref{op:rankGrasps}) by first averaging the quality 
metric of each grasp across all samples and then rank the grasps according to 
the new average quality metric, where the highest ranked grasp corresponds to 
highest average quality metric. It follows that the highest ranked grasp is 
also the solution to \eqref{eq:graspMax} which, in this work, is considered the 
most robust grasp over the shape uncertainty. 

\begin{algorithm}[t]
	\caption{\label{alg:grasp_planning}Robust Grasp Planning Over Uncertain Shape}
	\begin{algorithmic}[1]
		\STATE \textbf{Inputs:} Point-cloud $\matr{P}$ and number of dropout samples $I$
		\STATE Initialize empty object voxel set $O_{\mathcal{I}}\leftarrow \left\lbrace\right\rbrace$,		
		\STATE Initialize empty object mesh set $O_{\mathcal{M}}\leftarrow \left\lbrace\right\rbrace$,
		\STATE Initialize empty grasp sets $G, 
		\hat{G}\leftarrow\left\lbrace\right\rbrace$,
		\STATE Initialize empty grasp quality metric set $S\leftarrow\left\lbrace\right\rbrace$
		\STATE  $\matr{P}_{V}\leftarrow\textsc{Voxelize}(\matr{P})$\label{op:voxelize}	
		\FORALL{$i=1,\dots,I$} \label{op:dropoutsamplesbegin}
		\STATE Sample dropout mask $B$ \label{op:dropoutsample}
		\STATE $o_i\leftarrow\mathcal{H}_B(\matr{P}_V)$ \label{op:createsample}
		\STATE $o_m \leftarrow\textsc{ShapeCompletion}(o_i,\matr{P})$ \label{op:createMesh}
		\STATE $O_{\mathcal{I}} \leftarrow O_{\mathcal{I}} + \left\lbrace o_i\right\rbrace$ \label{op:addVoxel}
		\STATE $O_{\mathcal{M}} \leftarrow O_{\mathcal{M}} + \left\lbrace o_m\right\rbrace$ \label{op:addMesh}
		\ENDFOR \label{op:dropoutsamplesend}
		\STATE $\hat{O}\leftarrow \text{E}\left[O_{\mathcal{I}}\right]$\label{op:meanvoxel} 
		\STATE $\hat{O}\leftarrow\textsc{ShapeCompletion}(\hat{O},\matr{P})$ \label{op:meanmesh}
		\STATE $\hat{G}\leftarrow \textsc{PlanGrasp}(\hat{O})$ 
		\label{op:graspPlanning}
		\FORALL{$\hat{g} \in \hat{G}$} \label{op:robustPlanningBegin}
			\STATE  $S_g\leftarrow\left\lbrace\right\rbrace$
			\FORALL{$o_m \in O_{\mathcal{M}}$}
				\STATE $s \leftarrow \textsc{EvaluateGrasp}(\hat{g},o_m)$ 
				\label{op:evalGrasp}
				\STATE $S_g \leftarrow S_g + \left\lbrace s\right\rbrace$ \label{op:addGraspQuality}
			\ENDFOR
		\STATE $S \leftarrow S + \left\lbrace S_g\right\rbrace$ \label{op:addGraspQualities}
		\ENDFOR \label{op:robustPlanningEnd}
		\STATE $G \leftarrow 
		\textsc{RankGrasps}(\hat{G},S)$\label{op:rankGrasps}
		\RETURN $G$ \label{op:return}
	\end{algorithmic}
\end{algorithm}

%% file: sections/exp.tex
\section{Experiments}
\label{sec:exp}
The two main questions we wanted to answer in the experiments were:
\begin{enumerate}
	\item What is the shape reconstruction accuracy of the proposed method? 
	\item What is the impact of estimating shape distributions on grasp quality 
	and grasp success rate? 
\end{enumerate}

In order to provide justified answers to these questions we conducted three 
separate experiments. The first experiment (\secref{sec:GCR}) examines general 
shape reconstruction accuracy, the second one (\secref{sec:GSim}) evaluates grasp quality 
metrics in simulation, while the third experiment (\secref{sec:GReal}) evaluates grasp 
success rate on real hardware. In the first two experiments we compare our 
method with dropout sampling, ours without dropout sampling\footnote{by without 
dropout sampling we mean that dropout layers were enabled during training, but 
then disabled at test-time.}, and Varley's method \cite{varley_shape_2017} which is the only shape completion method proposed for grasping. In 
the third experiment we only compare our method with dropout sampling to 
Varley's. Henceforth, we refer to the three methods as \ac{ods}, \ac{od}, and 
\ac{va} respectively.

\subsection{Experimental Setup}
\label{sec:exp_setup}
For training and testing the network proposed in \secref{sec:SBSC} we used the 
same data as in \cite{varley_shape_2017}, that is voxelized occupancy grids of 
objects from the YCB and Grasp Database. The test-data consisted of two 
separate sets, holdout views and holdout model: the former is novel views of 
the objects used for training, while the latter are completely novel objects. 
The network itself was implemented in PyTorch 0.3.0 with a dropout rate of 0.2, 
and trained with \ac{adam}\cite{kingma2014adam} using a batch size of 32 and 
for 181 epochs\footnote{Code available at: 
\href{http://irobotics.aalto.fi/software-and-data/shape-completion}{irobotics.aalto.fi/software-and-data/shape-completion}}.
 The training was carried out on an NVIDIA Titan Xp and lasted for 
approximately a week. For evaluating \ac{va} we used a pre-trained network made 
publicly available by the 
authors\footnote{\href{http://shapecompletiongrasping.cs.columbia.edu/}{shapecompletiongrasping.cs.columbia.edu}}.

In the first two experiments we generated test data with the same procedure as 
in \cite{varley_shape_2017}, that is randomly sampling 50 
views from the training set (\emph{Training Views}), 50 views from the holdout 
view set (\emph{Holdout Views}), and  50  views  from the holdout  models set 
(\emph{Holdout  Models}). In the real world experiments the methods were 
evaluated on the 10 objects shown in \figref{fig:real_objects}.

We used \graspit{} \cite{miller2004graspit} to generate grasp candidates in 
both the simulated and real world grasping experiments. As our method is 
agnostic to the type of quality metric, in simulation we decided to evaluate 
two different ones: $\epsilon$- and the $v$-measure \cite{miller1999examples} 
as the former represents the quality metric of a worst case grasp and the 
latter an average case grasp. More specifically, the $\epsilon$-measure 
represents the radius of the largest 6D ball centered at the origin that can be 
enclosed by the convex hull of the wrench space, while the $v$-measure 
represents the volume of that convex hull. On the real hardware, however, only 
the $\epsilon$-measure was used as it attained a higher grasp success rate for both methods, 
according to a small pilot.

\subsection{General Completion Results}
\label{sec:GCR}
To evaluate the general shape completion results we use the Jaccard similarity, 
which is defined as
\begin{align}\label{eq:jaccard}
J(A,B)=\frac{\left|A\cap B\right|}{\left|A\cup B\right|}\enspace.
\end{align}
where $A$ and $B$ are two sets.
In this work $A$ is the ground truth and $B$ is the shape reconstructed with shape completion. In order to generate the sets $A$ and $B$ we follow the same procedure as in \cite{varley_shape_2017}, that is to voxelize each mesh to a resolution of $40^3$. 

To quantify the reconstruction results for \ac{ods} we generated 10 dropout 
samples and evaluated the Jaccard similarity on the mean mesh. The shape 
reconstruction results for \ac{va}, \ac{od}, and \ac{ods} are presented in 
\tabref{tb:jac_results}. We can see that there is essentially no difference 
between the three methods, meaning that our network---with and without dropout 
enabled at test-time---scored equally well as Varley's, a network that is 
approximately 10 times larger than ours. Moreover, the results do indicate that 
the mean shape approximated by averaging shape samples is representative for 
the mean of the unknown underlying shape distribution.

It is worth pointing out that the presented Jaccard similarity are 
substantially lower than the ones reported in \cite{varley_shape_2017} where 
\ac{va} achieves a reported score of 0.7771, 0.7486, and 0.6496, on training 
views, holdout views, and holdout models respectively. One reason for not 
attaining similar scores here is that we do not know the exact objects they 
included in the different test sets. Instead, we 
had to sample new ones and because of that some object that were in the test 
sets in \cite{varley_shape_2017} were most likely not in the test sets here and 
vice versa.

\begin{table}
	\centering
	\ra{1.2}\tbs{19}
	\caption{\label{tb:jac_results}Jaccard similarity results (higher is 
	better)}
	\begin{tabular}{@{}llll@{}}
		\toprule
		View & \multicolumn{1}{c}{\ac{va}} & \multicolumn{1}{c}{\ac{od}} & 
		\multicolumn{1}{c}{\ac{ods}} \\
		\midrule
		Training Views & 0.6205 & \textbf{0.6480} & 0.6446 \\
		Holdout Views & 0.6143 & 0.6382 & \textbf{0.6389} \\
		Holdout Models & 0.5632 & 0.5573 & \textbf{0.5651} \\
		\bottomrule
	\end{tabular}
\vspace{-0.3cm}
\end{table}

\subsection{Grasping in Simulation}
\label{sec:GSim}
In the simulated grasping experiment we evaluated the methods' capability at 
recognizing good grasp candidates. We evaluated 600 grasps for each of the 150 
shape completed objects in the test sets. To quantitatively compare grasps 
between methods, grasp directions were uniformly sampled around the object 
using \graspit{}. Then the two separate grasps for each method, one achieving 
the highest $\epsilon$-measures and the other the highest $v$-measures, were 
evaluated on the ground truth object.

For the last step we followed the same strategy as in \cite{varley_shape_2017}, 
that is to swap the shape completed object for the ground truth in \graspit{}, 
place the hand 20cm backward along the grasp approach vector, set the spread of 
its fingers, move it along the grasp approach vector until the pose was reached, 
and finally close the fingers.

The methods compared were \ac{va}, \ac{od}, and \ac{ods}. \ac{ods} was 
evaluated on 10 dropout samples using the grasp planning method detailed in 
\secref{sec:RobustGraspPlanning}. The other methods chose the grasps according 
to their performance on the point estimate of the shape. 

To analyze the statistical differences between the methods we used a one sided 
Wilcoxon signed-rank test; the results are presented in 
\tabref{tb:sim_results}. Based on these results we can draw several interesting 
conclusions. For once, there is a statistical significant improvement using 
\ac{ods} over \ac{va} for determining the most robust grasp in terms of both 
$\epsilon$- and $v$-measure. A similar statistical significant improvement was 
also visible for \ac{ods} over \ac{od} but only in terms of $v$-measure. 
The reason for \ac{ods} outperforming both \ac{va} and \ac{od} is 
that the performance of these methods deteriorates heavily in the shift from 
training or holdout views to holdout models, indicating that they are not able 
to recognize high quality grasps on novel objects. For instance, the relative 
performance drop for \ac{va} and \ac{od} from training views to holdout models 
are -51.5\% and -60\% for $\epsilon$-measures and -54.1\% and -16.7\% for 
$v$-measures, respectively. On the other hand \ac{ods} loses only -4.5\% for 
$\epsilon$-measure and gains +21\% for $v$-measures, that is \ac{ods} actually 
performs better on novel objects in terms of the $v$-measure. These results 
demonstrate the importance of including shape uncertainty in grasp 
planning especially in cases where the uncertainty is higher, such as with  novel 
objects.

\begin{table*}
	\centering
	\ra{1.4}
	\caption{\label{tb:sim_results}Average $\epsilon$ and $v$-quality metrics 
	over different test sets, with test statistics and $p$-values of pair-wise 
	one sided Wilcoxon signed-rank test for \ac{va} vs. \ac{ods} and \ac{od} vs. \ac{ods}. \ac{ods} was evaluated 
	on 10 dropout samples.}
		\begin{tabular}{@{}lllllllllllll@{}}
		\toprule
		& \multicolumn{2}{c}{\ac{va}} & \multicolumn{2}{c}{\ac{od}} & 
		\multicolumn{2}{c}{\ac{ods}} & \multicolumn{2}{c}{\ac{va} vs \ac{ods}} 
		& \multicolumn{2}{c}{\ac{od} vs \ac{ods}} \\
		\cmidrule(lr){2-3}\cmidrule(lr){4-5}\cmidrule(lr){6-7}\cmidrule(lr){8-9}\cmidrule(lr){10-11}
		& \epst{} & \vt{} & \epst{} & \vt{} & \epst{} & \vt{} & \epst{} & \vt{} 
		& \epst{} & \vt{} \\
		\midrule
		Training Views & 0.0594 & 0.1212 & 0.0757 & 0.1674 & 0.0682 & 0.1976 & 
		\nat{} & \nat{} & \nat{} & \nat{} \\
		Holdout Views & 0.0614 & 0.1504 & 0.0590 & 0.1297 & 0.0584 & 0.2127 & 
		\nat{} & \nat{} &\nat{} & \nat{} \\
		Holdout Models & 0.0288 & 0.0556 & 0.0300 & 0.1394 & 0.0651 & 0.2404 & 
		\multicolumn{1}{l}{T=102, 
		p\textless{}.001***\cellcolor{verysignificant}} & 
		\multicolumn{1}{l}{T=42, 
		p\textless{}.001***\cellcolor{verysignificant}} & 
		\multicolumn{1}{l}{T=167.5, 
		p\textless{}.01**\cellcolor{verysignificant}} & \nat{} \\ 
		\bottomrule
	\end{tabular}
\end{table*}

\subsection{Grasping on Real Hardware}
\label{sec:GReal}
\begin{figure}
	\centering
	\includegraphics[width=.9\linewidth]{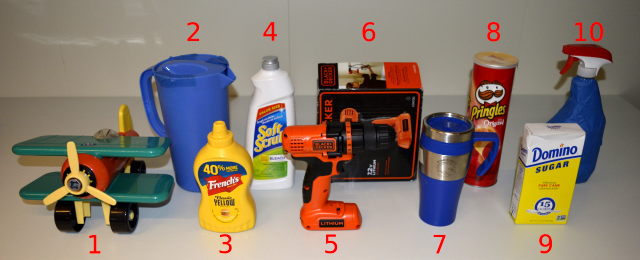}
	\caption{\label{fig:real_objects}The 10 different objects with their 
	corresponding number. All objects except number 7 are from the YCB object 
	set.}
\vspace{-0.5cm}
\end{figure}

As a final experiment we compared \ac{ods} with 20 dropout samples and \ac{va} in terms of grasp 
success rate on real hardware. To this end, we used a \barrett{} mounted on a 
\panda{} to grasp the 10 different objects visualized in 
\figref{fig:real_objects}. Out of these objects 1, 2, 4, 5, 7, and 10 have one 
axis of symmetry; 3, 6 and 9 have two; and 8 has three.

We ran the complete grasping pipeline for each object in five different 
orientations (0\textdegree, 72\textdegree, 144\textdegree, 216\textdegree, and 
288\textdegree) and from two different camera viewpoints: one looking at the 
object from the left side of the robot and the other from the opposite side of 
the object to the robot. In total this setup amounts to 100 grasps per method. 
\graspit{}'s simulated annealing planner \cite{ciocarlie2007dimensionality} was 
used to plan and evaluate grasps as this sped up the process of planning and 
evaluating grasps compared to the uniform planning process used in 
\secref{sec:GSim}. 

To evaluate if a grasp was force-closure, the robot moved to the planned grasp 
pose, then closed the hand and moved the arm upward 20cm, then moved back to 
the starting position and finally rotated the hand $\pm$90\textdegree{} around the 
last joint. If the object was stable in the hand for this whole procedure we 
deemed it force-closure. If the robot, on the other hand, was not able to grasp 
the object or the object moved inside the hand during the arm motion we deemed 
it not force-closure.

\begin{figure}
	\centering
	\includegraphics[width=0.9\linewidth]{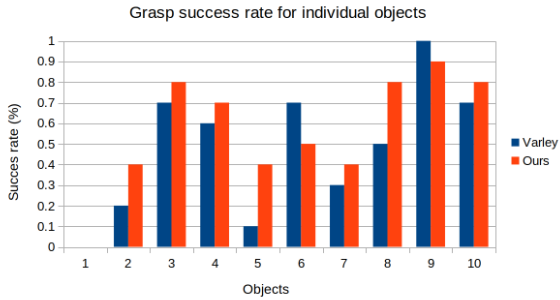}
	\caption{\label{fig:successrate_real_objects}The individual grasp success 
	rate for both methods on each of the 10 objects used in the experiment.}
\vspace{-0.2cm}
\end{figure}

\begin{table}
	\centering
	\ra{1.3}\tbs{20}
	\caption{\label{tb:real_exp_summary}Real Hardware experiment results. 
	\ac{ods} was evaluated on 20 dropout samples.}
	\begin{tabular}{@{}lll@{}}
		\toprule
		& \ac{va} & \ac{ods} \\
		\midrule
		Grasp Success Rate (\%) & 48 & \textbf{59} \\
		Shape Completion Time (s) & \textbf{7.4}   & 86.73\\
		Grasp Evaluation Time (s) & \textbf{16.53} & 83.25\\ 
		\bottomrule                                                         
	\end{tabular}
\vspace{-0.1cm}
\end{table}

\begin{figure*}%
	\centering
	\begin{subfigure}[b]{.28\columnwidth}
		\includegraphics[width=0.9\columnwidth]{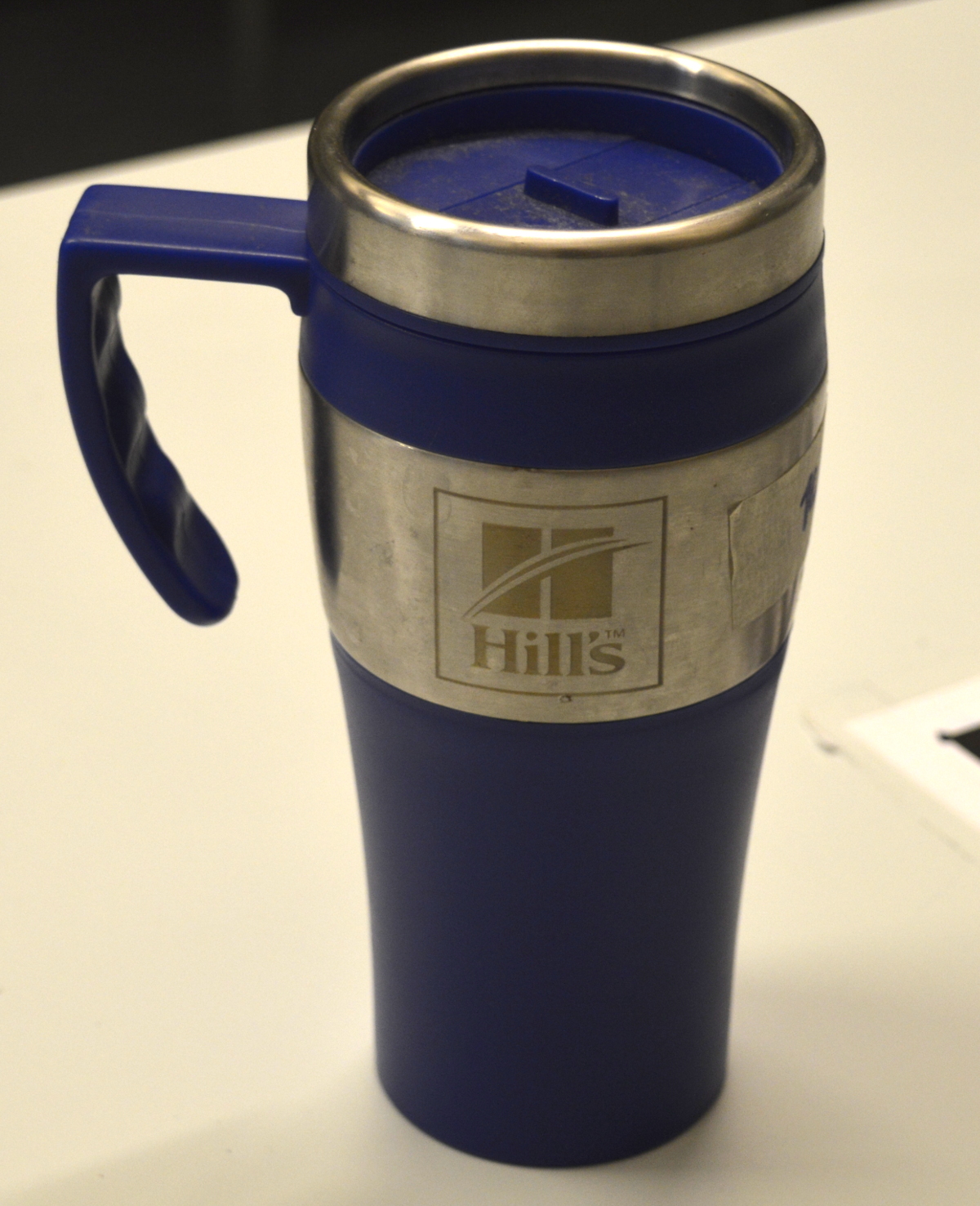}%
		\caption{\label{fig:sub_a}}	
	\end{subfigure}
	\begin{subfigure}[b]{.28\columnwidth}
		\includegraphics[width=0.9\columnwidth]{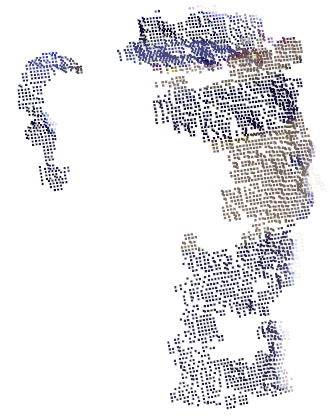}%
		\caption{\label{fig:sub_b}}	
	\end{subfigure}
	\begin{subfigure}[b]{.28\columnwidth}
		\includegraphics[width=1\columnwidth]{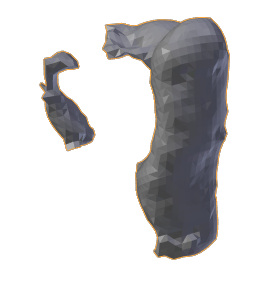}%
		\caption{\label{fig:sub_c}}	
	\end{subfigure}
	\begin{subfigure}[b]{.28\columnwidth}
		\includegraphics[width=1\columnwidth]{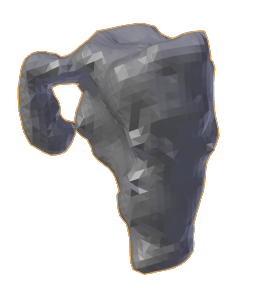}%
		\caption{\label{fig:sub_d}}	
	\end{subfigure}
	\begin{subfigure}[b]{.28\columnwidth}
		\includegraphics[width=1\columnwidth]{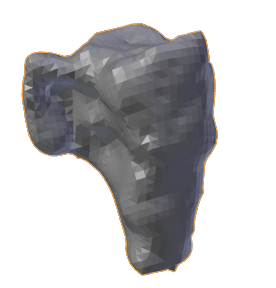}%
		\caption{\label{fig:sub_e}}	
	\end{subfigure}
	\begin{subfigure}[b]{.28\columnwidth}
		\includegraphics[width=1\columnwidth]{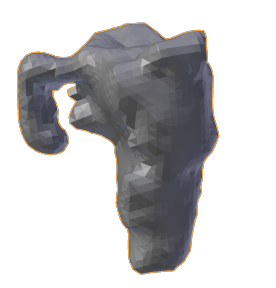}%
		\caption{\label{fig:sub_f}}	
	\end{subfigure}
	\begin{subfigure}[b]{.28\columnwidth}
		\includegraphics[width=1\columnwidth]{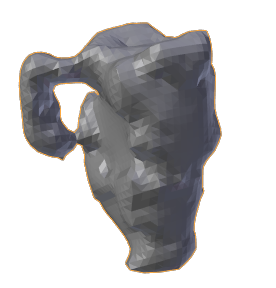}%
		\caption{\label{fig:sub_g}}	
	\end{subfigure}
	\caption{\label{fig:shape_completed_mug}\subfigref{fig:sub_a} shows the 
	real object and a \pc{} of it is visible in \subfigref{fig:sub_b}. Given 
	this \pc{}, \subfigref{fig:sub_c} shows the shape completed object using 
	\ac{va}	and \subfigref{fig:sub_d} the mean shape completed object using 
	\ac{ods}. Three 
	of the in total twenty samples used to generate \subfigref{fig:sub_d} are 
	shown in \subfigref{fig:sub_e}-\subfigref{fig:sub_g}.}
\end{figure*}

The experimental results are shown in \tabref{tb:real_exp_summary}, where we
report the percentage of successful grasp attempts (\emph{Gasp Success Rate}), 
the average time a method required to complete a mesh from a partial view 
(\emph{Shape Completion Time}), and the average time the planner required to 
plan and evaluate individual grasps (\emph{Grasp Evaluation Time}). To analyze 
the statistical differences between the methods in terms of grasp success rate 
we used a one sided Wilcoxon signed-rank test. The test showed a statistical 
significant improvement (T=203.5, p\textless{}.05*) on grasp success rate of 
\ac{ods} over \ac{va}. 
Furthermore, \figref{fig:successrate_real_objects} shows that the grasping 
performance on individual shapes varies a lot between methods. For example, no 
method managed to generate a stable grasp for object 1 (the toy airplane) due 
to its low frictional surface. If we do not consider object 1, our method was 
more robust on grasping objects with only one axis of symmetry (objects 2, 5, 7 
and 10), whereas for the objects with more axis of symmetry \ac{va} and 
\ac{ods} were better on two objects each, observations that may result from 
random effects. 
Together, based on the above results, it stem to reason that including shape 
uncertainty when planning grasps compared to only planning on a point estimate 
of the shape improves grasps success rate especially on complex objects which 
are more difficult to complete.


Again, as was noted in \secref{sec:GCR}, we did not achieve similar results for 
\ac{va} as reported in \cite{varley_shape_2017} where the grasp success rate 
was 93.33\%. This difference is most likely due to the fact that we perform 
many more grasps (100 compared to 15) and use fewer easy-to-complete shapes 
such as boxes (1 compared to 4) and instead included harder objects, \eg{} the 
metallic cup (see \figref{fig:shape_completed_mug}).

In \figref{fig:shape_completed_mug} we show one shape completion example. It is 
clearly visible that from the \pc{} in \figref{fig:sub_b} the shape completed 
object using \ac{va} (\figref{fig:sub_c}) severely underestimates the thickness 
of the object and is unable to connect the handle to the body of the cup. In 
contrast the mean object shape using \ac{ods} (\figref{fig:sub_d}), although 
still far from perfect, is definitely better at estimating the real thickness 
of the cup and is also able to connect the handle to it. Three of the twenty 
samples used to create the mean mesh are visualized in 
\figsref{fig:sub_e}{fig:sub_g} and individually they show very interesting 
behaviors. For example, the object shown in \figref{fig:sub_e} is very thick 
and completely fills the cup while the one in \figref{fig:sub_f} is rather thin 
in the bottom half. The object in \figref{fig:sub_g}, on the other hand, models 
the overall shape well but is instead irregular. Viewed together, the samples 
are consistent in areas covered by the \pc{}, which is expected as the 
confidence there is high, while more uncertain in areas that are occluded to 
the \pc{}. As the different samples capture different possible shape completed 
object, planning grasps that are good on all samples makes the grasps more 
robust to shape uncertainty.

Although our method achieved a higher grasp success rate it also requires 
longer computational time as seen in \tabref{tb:real_exp_summary}. It is, 
however, possible to substantially lower both the completion and evaluation 
time by harnessing the inherent parallel structure of 
\algoref{alg:grasp_planning}. For instance, to lower evaluation time all grasps 
could be evaluated in parallel on each of the object samples. Similarly, the 
shape completion time could be lowered by doing shape sampling in parallel 
through multiple copies of the \ac{ods} network with individual dropout masks. 



%% file: sections/concl.tex
\section{Conclusions}
\label{sec:concl}
We presented a method for generating robust grasps over uncertain shape 
completions. The key insight was to use dropout layers not only during training 
but also at run-time to generate shape samples and then rank grasps based on 
their joint quality metrics over all the samples. We compared our method to 
current \sota{} shape completion methods used in robotics both in simulation 
and on real hardware. Together all results from shape reconstruction, 
simulation, and real hardware indicated that including shape uncertainty did 
lead to statistical significant improvements in terms of recognizing good 
grasps and achieving higher grasp success rate while keeping shape 
reconstruction quality comparable to the benchmark. 

In conclusion the work presented here demonstrates that planning 
shape-uncertainty-aware grasps brings significant advantages over solely 
planning on a good point estimate. This, in turn, poses new interesting research 
questions. For instance, can more refined shape completion networks 
\cite{han_high-resolution_2017,yang20173d} benefit from modeling uncertainty? 
Similarly, from a robotics perspective, can the performance of end-to-end 
methods such as Dex-Net \cite{mahler2017dex} improve if they also included 
uncertainty as a part of the network? These questions pave the way for 
interesting future research avenues.